\documentclass{article}
\usepackage{spconf,amsmath,graphicx,hyperref}
\usepackage{multirow}
\usepackage{amssymb}
\usepackage{algorithm}    
\usepackage{algpseudocode}

\title{Inverse Rendering for High-Genus 3D Surface Meshes from Multi-view Images with Persistent Homology Priors}
\name{Xiang Gao\textsuperscript{1}, 
Xinmu Wang\textsuperscript{1}, 
Yuanpeng Liu \textsuperscript{1},
Yue Wang\textsuperscript{1},
Junqi Huang\textsuperscript{2}, 
Wei Chen\thanks{* Corresponding author.}\textsuperscript{3}\textsuperscript{*}, 
Xianfeng Gu\textsuperscript{1}}

\address{%
\textsuperscript{1}Department of Computer Science, Stony Brook University, Stony Brook, NY, USA\\
\textsuperscript{2}Department of Applied Mathematics \& Statistics, Stony Brook University, Stony Brook, NY, USA\\
\textsuperscript{3}Academy for Multidisciplinary Studies, Capital Normal University, Beijing, China\\
\normalsize \{gao2, xinmuwang, yuanpliu, wang139, gu\}@cs.stonybrook.edu, 
junqi.huang@stonybrook.edu, chen@cnu.edu.cn
}
\begin{document}
%
\maketitle

\begin{abstract}
Reconstructing 3D objects from images is inherently an ill-posed problem due to ambiguities in geometry, appearance, and topology. This paper introduces collaborative inverse rendering with persistent homology priors, a novel strategy that leverages topological constraints to resolve these ambiguities. By incorporating priors that capture critical features such as tunnel loops and handle loops, our approach directly addresses the difficulty of reconstructing high-genus surfaces. The collaboration between photometric consistency from multi-view images and homology-based guidance enables recovery of complex high-genus geometry while circumventing catastrophic failures such as collapsing tunnels or losing high-genus structure. Instead of neural networks, our method relies on gradient-based optimization within a mesh-based inverse rendering framework to highlight the role of topological priors. Experimental results show that incorporating persistent homology priors leads to lower Chamfer Distance (CD) and higher Volume IoU compared to state-of-the-art mesh-based methods, demonstrating improved geometric accuracy and robustness against topological failure.

\end{abstract}
\section{Introduction}

Reconstructing high-genus 3D surface meshes from images arises in applications such as VR/AR, CAD, 3D printing, and scientific simulation or visualization, where preserving complex topological structures is crucial.  While low-genus (genus-0) meshes can often be recovered from multi-view images, high-genus surfaces remain challenging due to topological ambiguity, which frequently causes tunnels and handles to collapse or disappear, leading to reconstructions with incorrect topology.
Within a 3D surface mesh-based inverse rendering framework, the goal is to optimize the mesh so that its rendered appearance matches the input images.
Existing studies~\cite{Nicolet2021Large, Jung2023, gao2025inverserenderinghighgenussurface} have attempted to reconstruct 3D surface meshes from multiview images using gradient-based optimization such as Adam \cite{KingBa15}. Nicolet et al.\cite{Nicolet2021Large} introduced an inverse rendering framework that enables accurate mesh reconstruction for genus-0 surfaces. Inspired by this framework, Jung et al.\cite{Jung2023} extended it with a template-based approach and adaptive mesh density in feature-rich regions. Although these approaches achieve impressive 3D mesh reconstructions on low-genus surfaces, they fall short on complex high-genus structures, limiting their suitability for real-world applications.
In addition, the standard practice of adopting uniform camera placement on a sphere is inherently biased and unjustified, as it assumes simple closed genus-0 geometries homeomorphic to a sphere. The rendered multiviews often produce vanishing or exploding gradients near tunnel regions, which appear as hole-like structures in the images. Moreover, severe occlusions across views can yield conflicting gradients, causing the optimization to compete across different viewpoints and ultimately leading to topological fragmentation in the reconstruction. Even when a 3D surface mesh is available, it is not straightforward to determine optimal camera placements, which further complicates robust reconstruction. This biased and unjustified strategy has been widely adopted in non-mesh-based deep generative image-to-3D studies~\cite{liu2023zero1to3, guo2024tetsphere, long2023wonder3d, liu2023syncdreamer, otTalk, gao2025neural}. 

Persistent Homology (PH)~\cite{PH}, a powerful tool deeply rooted in algebraic topology, provides a principled way to detect topological invariants such as tunnel loops, offering a more explainable approach to place cameras.
This paper proposes a collaborative inverse rendering strategy that incorporates persistent homology priors with a widely adopted uniform-sphere assumption to provide a justified basis for camera placements. The key insight is to place cameras near topological tunnels, which are topological invariants detected by persistent homology, thereby capturing regions prone to gradient vanishing, occlusion, or topological collapse. Compared to the state-of-the-art mesh-based inverse rendering method of Nicolet et al.~\cite{Nicolet2021Large}, our strategy shows stronger robustness against topological failure, achieving lower Chamfer Distance (CD) and higher Volume IoU.

\begin{figure*}[t] 
    \centering
    \includegraphics[width=0.95\textwidth]{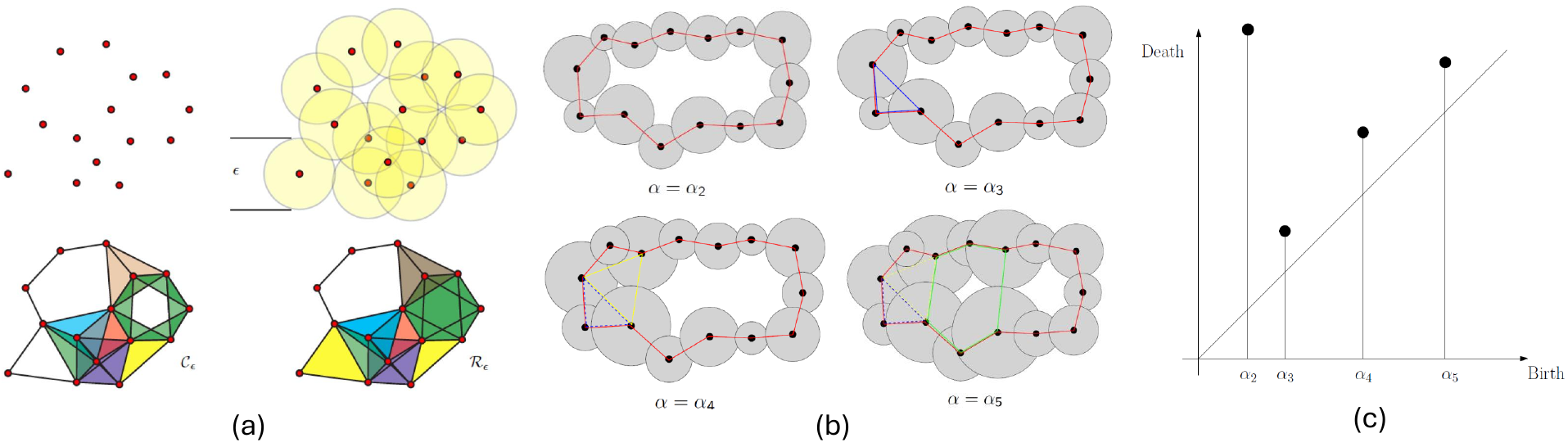} 
    \caption{The pipeline of computing a persistent diagram. (a). Comparison between \v{C}ech Complex \( \mathcal{C}_\varepsilon \) and Vietoris-Rips Complex \( \mathcal{R}_\varepsilon \) \cite{ghrist2008barcodes}. (b). The evolution of filtered Cech complexes with varying scale parameters. (c). Persistent Diagram.}
    \label{fig:overviewl}
\end{figure*}

\section{Methodology}
\subsection{Problem Formulation}
We formulate the inverse rendering problem for high genus surface meshes as:

\begin{equation} 
\begin{aligned}
   \arg\min_{\mathbf{x}} \quad & \Phi(R(\mathbf{x})) + \mathbf{w_1}\operatorname{tr}(\mathbf{x}^T \mathbf{L} \mathbf{x}) \\
    \text{s.t.} \quad & \det(\mathbf{J}_R^{(k)}) > 0, \quad \forall k \in \{1, \dots, |F|\}
\end{aligned}
\label{eq:constrained}
\end{equation} 
where \( \mathbf{x} \in \mathbb{R}^{n \times 3} \) are mesh vertex positions, \( \mathbf{L} \in \mathbb{R}^{n \times n} \) the uniform bi-Laplacian \cite{Botsch2004a, Jacobson2010}, \( R(\cdot) \) the rendering function, \( \Phi(\cdot) \) the rendering loss, \( \operatorname{tr}(\cdot) \) the matrix trace, \( |F| \) the number of faces, and \( \mathbf{J}_R^{(k)} \) the Jacobian of triangle \( k \). Reformulating Equation~\ref{eq:constrained} as an unconstrained optimization gives:

\begin{equation} 
\begin{aligned}
    \arg\min_{\mathbf{x}} \, &\Phi(R(\mathbf{x})) + \mathbf{w_1} \operatorname{tr}(\mathbf{x}^T \mathbf{L} \mathbf{x}) \\
    &+ \mathbf{w_2} \sum_{k=1}^{|F|} \left( \min \{ 0, \det(\mathbf{J}_R^{(k)}) \} \right)^2
\end{aligned}
\label{eq:unconstrained}
\end{equation} 
which can be optimized by gradient descent, with \( \mathbf{w_1} \) enforcing smoothness and \( \mathbf{w_2} \) penalizing triangle inversion.

\subsection{Preliminary on Persistent Homology (PH)}
Persistent homology is a central tool in Topological Data Analysis (TDA) for tracking the evolution of topological features across scales~\cite{Edelsbrunner2002,EdelsbrunnerHarer2022}. A simplicial complex \( \mathbb{K} \) is a collection of simplices closed under inclusion and intersection, where a \( k \)-simplex \(\sigma^k = [x_0,\dots,x_k]\) is the convex hull of \(k+1\) affinely independent points, such as vertices (\(0\)-simplices), edges (\(1\)-simplices), triangles (\(2\)-simplices), and tetrahedra (\(3\)-simplices). For a point set \( \{x_\alpha\} \subset \mathbb{R}^n \), the \v{C}ech complex \( \mathcal{C}_\varepsilon \) includes a simplex if the corresponding closed \(\varepsilon/2\)-balls intersect, while the Vietoris-Rips complex \( \mathcal{R}_\varepsilon \) includes a simplex if all pairwise distances between its vertices are at most \(\varepsilon\). A filtration is then a nested sequence of complexes \(\emptyset = \mathbb{K}_{-1} \subset \mathbb{K}_0 \subset \cdots \subset \mathbb{K}_n = \mathbb{K}\), parameterized by \(\varepsilon\), that captures topological changes across scales. In this setting, a \(k\)-chain is a linear combination of \(k\)-simplices forming the group \(C_k(\mathbb{K})\), with the boundary operator \(\partial_k: C_k \to C_{k-1}\) satisfying \(\partial_{k-1}\circ \partial_k=0\). Closed chains (\(\partial_k \gamma=0\)) form the subgroup \(Z_k\), exact chains (\(\gamma=\partial_{k+1}\sigma\)) form \(B_k\), and the corresponding homology groups are defined as \(H_k(\mathbb{K}) = Z_k(\mathbb{K})/B_k(\mathbb{K}) = \ker \partial_k / \operatorname{im} \partial_{k+1}\), distinguishing cycles from boundaries. The inclusion maps of a filtration induce homomorphisms between homology groups, \(0 = H_p(\mathbb{K}_{-1}) \to H_p(\mathbb{K}_0) \to \cdots \to H_p(\mathbb{K}_n)\), thereby tracking the birth and death of features across scales. 

\subsection{Persistent Homology for Handle and Tunnel Loops}
\label{sec:ph}

\begin{figure}[htbh] 
    \centering
    \includegraphics[width=0.46\textwidth]{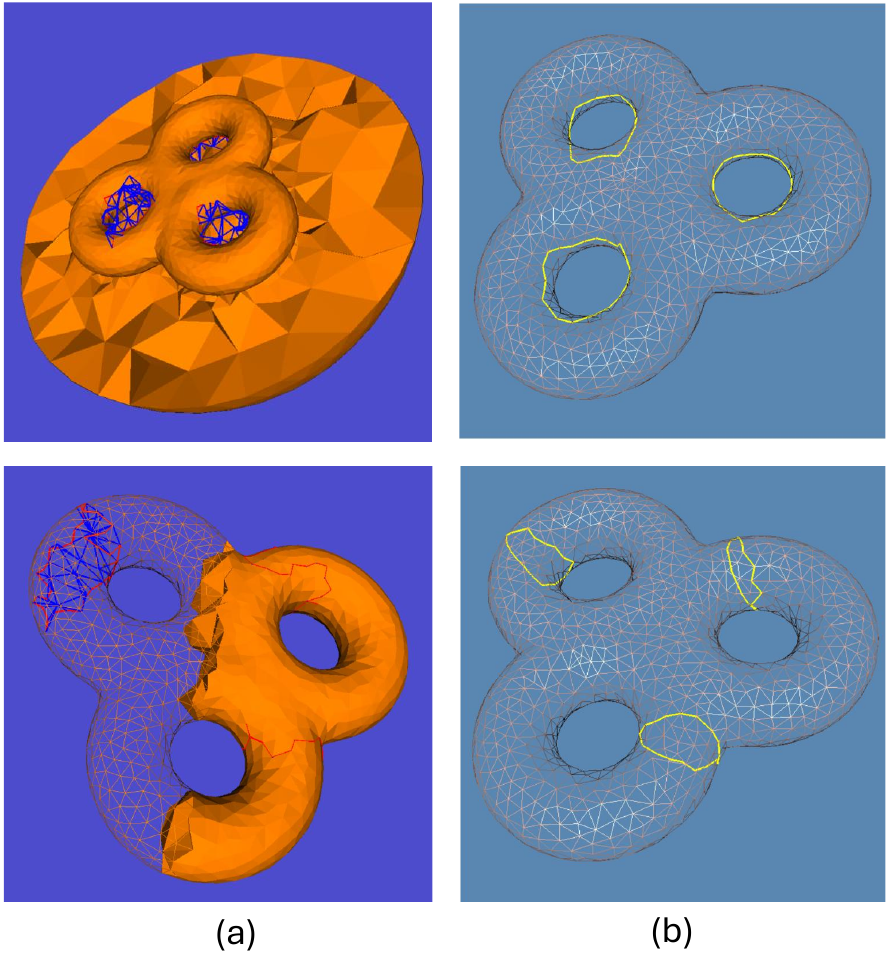} 
    \caption{Topological invariants captured by our implemented two-stage persistent homology algorithm on a torus-3 topology. (a) Exterior and interior volumetric meshes. (b) Handle and tunnel loops.}
    \label{fig:peristent}
\end{figure}

We compute handle and tunnel loops of closed surfaces using persistent homology. Given a filtration of simplicial complexes \(\emptyset = \mathbb{K}_{-1} \subset \mathbb{K}_0 \subset \cdots \subset \mathbb{K}_n = \mathbb{K}\), each simplex is classified as \emph{positive} if it creates a new homology class or \emph{negative} if it destroys one. A positive simplex gives rise to a non-exact cycle, while a negative simplex pairs with the youngest active generator in its boundary, ending its persistence interval. For a genus-\(g\) closed surface \(M\), the first homology group \(H_1(M)\) has rank \(2g\), so persistent homology yields \(2g\) unpaired edges as fundamental loops; adding interior simplices pairs \(g\) of them with interior faces to form handle loops, while adding exterior simplices pairs the other \(g\) with exterior faces to form tunnel loops. The full procedure, including the Pair Algorithm for generator–killer matching, the Mark Loop Algorithm to trace cycles, and Null Homological Cycle Detection to discard null-homologous loops, followed by Birkhoff curve shortening for refinement, is illustrated in Fig.~\ref{fig:overviewl}, where the computed handle and tunnel loops are shown in Fig.~\ref{fig:peristent}.

\begin{algorithm}[htbp]
  \caption{Pair Algorithm}
  \label{alg:pair}
  \begin{algorithmic}[1]
    \State $c \gets \partial_{p}\sigma$; $\tau \gets$ youngest positive $(p\!-\!1)$-simplex in $c$
    \While{$\tau$ is paired and $c \neq \emptyset$}
        \State find $(\tau,d)$, $d$ paired with $\tau$; $c \gets \partial_{p}d + c$
        \State update $\tau$ to youngest positive $(p\!-\!1)$-simplex in $c$
    \EndWhile
    \If{$c \neq \emptyset$} $\sigma$ is negative, paired with $\tau$\Else$\sigma$ is positive \EndIf
  \end{algorithmic}
\end{algorithm}

\begin{algorithm}[htbp]
  \caption{Mark Loop Algorithm}
  \label{alg:mark_loop}
  \begin{algorithmic}[1]
    \State $c \gets \partial_{2}d$; $\tau \gets$ youngest generator edge in $c$
    \While{$\tau$ is paired and $c \neq \emptyset$}
        \State find $(\tau,d)$; $c \gets \partial_{p}d + c$; update $\tau$
        \If{$\tau$ on boundary} \textbf{break} \EndIf
    \EndWhile
    \State \Return $c$
  \end{algorithmic}
\end{algorithm}

\begin{algorithm}[htbp]
  \caption{Null Homological Cycle Detection}
  \label{alg:null_homological_cycle_detection}
  \begin{algorithmic}[1]
    \State Build spanning tree $T$ of $G$; $G \setminus T = \{e_1,\dots,e_k\}$
    \State Construct $c_i = T \cup e_i$, $i=1,\dots,k$
    \State Compute persistent homology of $M$
    \State For each $c_i$, call NullDetection($c_i$)
  \end{algorithmic}
\end{algorithm}

\begin{figure}[htbh]
    \centering
    \includegraphics[width=1\columnwidth]{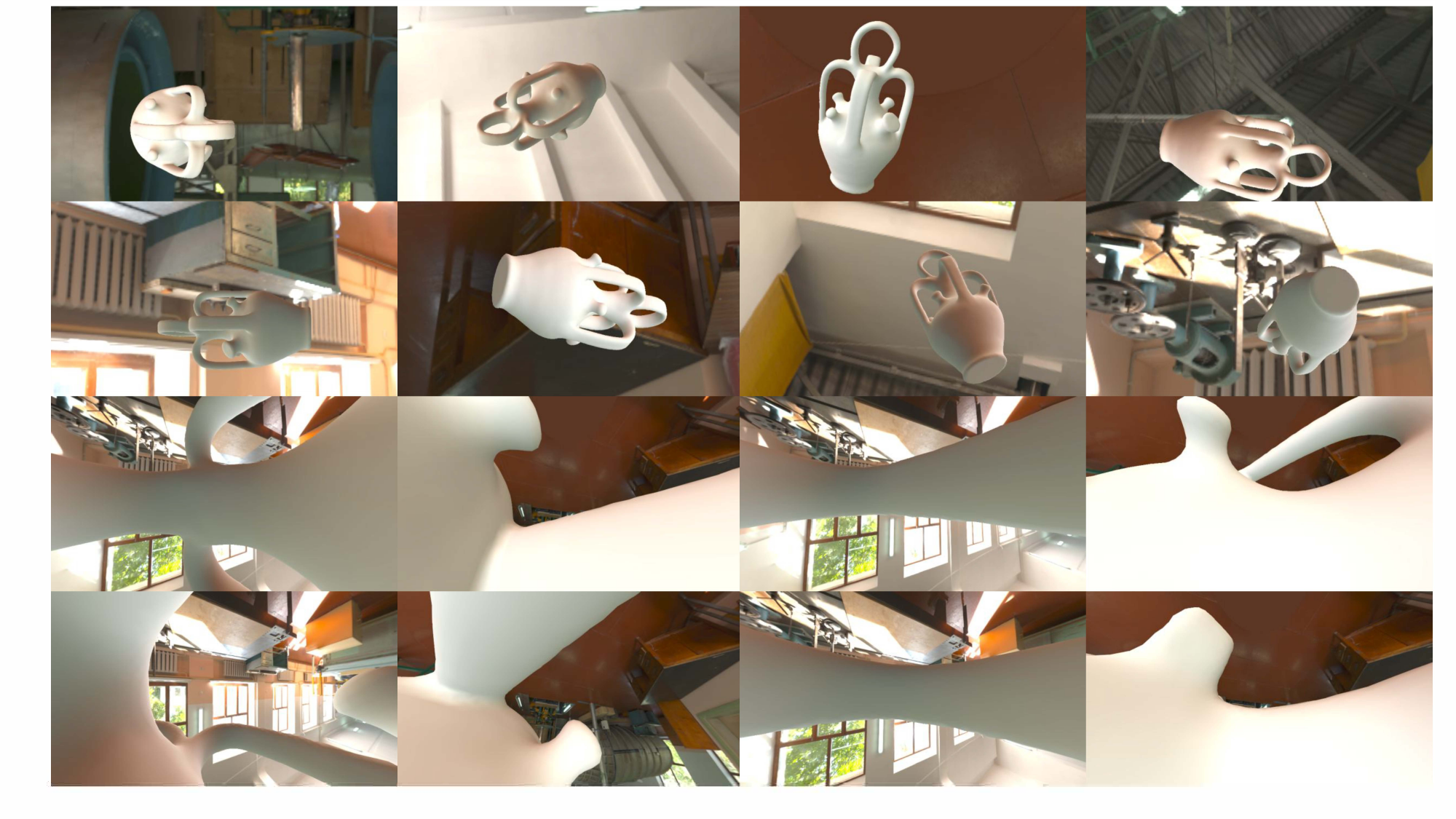}
    \caption{Collaborative rendering for the Botijo model. The top two rows show views obtained under the uniform-sphere assumption, while the bottom two rows show views guided by persistent homology priors.}
    \label{fig:collaborative_views}
\end{figure}

\section{Experiments and Results}

\noindent\textbf{Evaluation Dataset.} Our evaluation includes a total of eight relatively challenging high-genus models to test reconstruction accuracy. These models are selected from the GSO dataset~\cite{downs2022google} and the Thingi10K dataset~\cite{Thingi10K}, with the following characteristics. The models exhibit high-genus structures, and occlusions can occur in the multiview setting, posing challenges to mesh-based inverse rendering frameworks that rely solely on gradient-based optimization together with the biased uniform-sphere camera placement assumption.\\
\noindent\textbf{Baseline Methods.} Few existing studies explore neural network-free approaches for surface mesh reconstruction within inverse rendering frameworks, primarily due to the challenge of maintaining surface topology as the genus increases. To assess the effectiveness of our method, we compare it against the state-of-the-art approach by Nicolet et al.~\cite{Nicolet2021Large}, which relies on pure gradient-based optimization with the uniform-sphere camera placement assumption. We focus on this method as it is one of the few mesh-based approaches without neural networks and has recently been integrated into the Mitsuba3 library~\cite{Mitsuba3}, making it a solid baseline.

\subsection{Qualitative Results}
\begin{figure}[t] 
    \centering
    \includegraphics[width=0.49\textwidth]{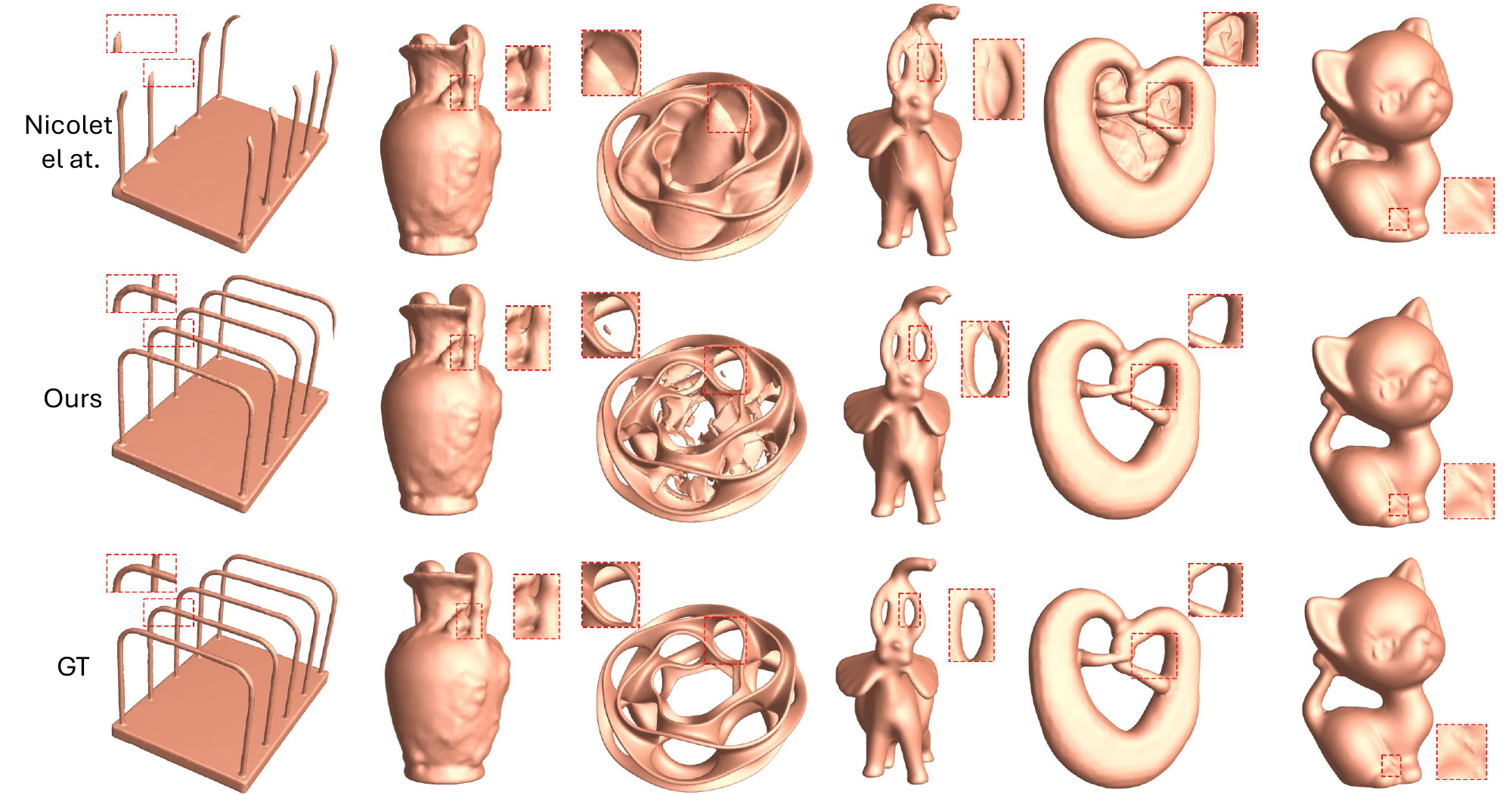} 
    \caption{Qualitative comparison on high-genus surface meshes with Nicolet et al.~\cite{Nicolet2021Large}. Our method, incorporating persistent homology priors, better preserves overall topological structures, particularly in regions with multiple tunnels.}
    \label{fig:qualitative_comparison}
\end{figure}

We present rendered views of reconstructed high-genus models using both Phong vertex and fragment shaders to facilitate a direct qualitative comparison with the baseline method. As illustrated in Figure~\ref{fig:qualitative_comparison}, the baseline approach exhibits several failure cases when reconstructing high-genus surfaces. In particular, the reliance on a uniform-sampling assumption for camera placement often leads to the collapse of critical topological features such as tunnels and handle loops, resulting in reconstructions with incorrect topology. In contrast, our method preserves these structures by explicitly incorporating persistent homology priors jointly with a collaborative inverse rendering strategy, rather than relying solely on uniform camera sampling, thereby producing reconstructions that maintain both the correct topology and geometric accuracy. As shown in Figure~\ref{fig:collaborative_views}, the top rows based on uniform sampling collapse tunnels and handles, whereas the bottom rows guided by persistent homology priors and collaborative inverse rendering preserve both topological structures and geometric detail. We find our reconstructions are closest to the ground truth because the proposed strategy is explicitly aware of tunnel and handle loops, ensuring that these features are better retained in the final geometry.

\begin{figure}[t] 
    \centering
    \includegraphics[width=\columnwidth]{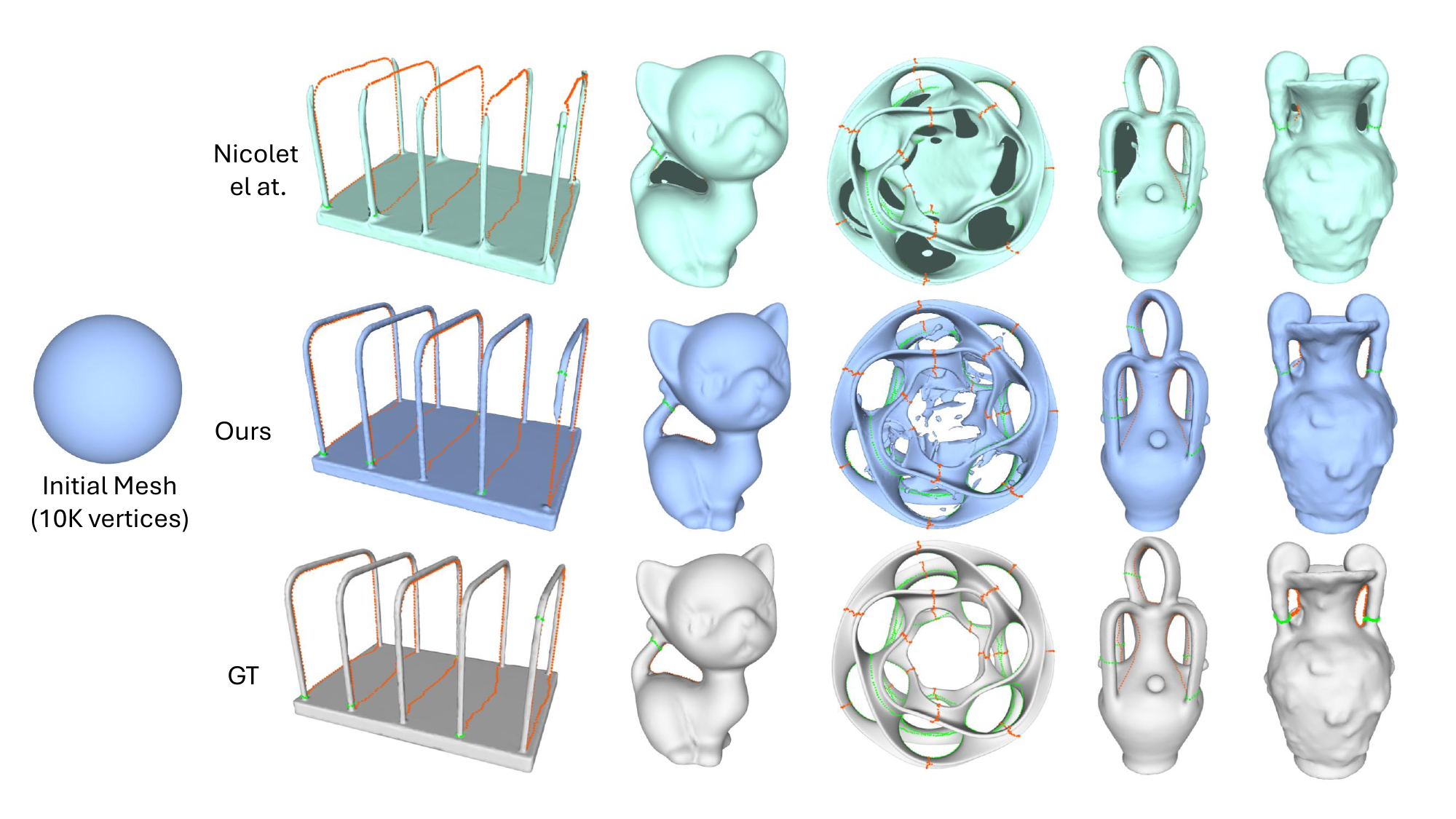} 
    \caption{Qualitative comparisons on high-genus surface meshes with tunnel loops (orange) and handle loops (green), compared to Nicolet et al.~\cite{Nicolet2021Large}.}
    \label{fig:teaser}
\end{figure}

In addition, we provide qualitative comparisons on high-genus surface meshes 
with topological annotations in Figure~\ref{fig:teaser}. Tunnel loops (orange) 
and handle loops (green) highlight critical structures often lost in baseline 
reconstructions (Nicolet et al.~\cite{Nicolet2021Large}). While the baseline 
struggles due to uniform camera sampling, our method preserves these features 
consistently across views, demonstrating both geometric fidelity and topological 
correctness, and confirming its effectiveness on complex high-genus models.

\subsection{Quantitative Results}
It follows from Table~\ref{tab:comparison} that the existing method of Nicolet et al.~\cite{Nicolet2021Large} provides competitive performance under certain conditions. However, as the table demonstrates, our approach consistently achieves lower reconstruction error and better preservation of topological structures. Importantly, the benchmark models considered here are high-genus surfaces, in contrast to simple genus-0 shapes such as the sphere. In this more challenging regime, the improvements are especially pronounced: by jointly leveraging persistent homology priors with the uniform-sampling assumption in camera placement, our method faithfully reconstructs multiple tunnel and handle loops, whereas prior work often suffers from topological collapse. These quantitative results clearly underscore the effectiveness of our approach in handling complex, high-genus surface geometries compared to baseline methods in this line of work.

\begin{table}[h]
    \centering
    \caption{Quantitative Results.}\vspace{1mm}
    \begin{tabular}{|c|c|c|c|c|}
        \hline
        \multirow{2}{*}{\textbf{Model}} & \multicolumn{2}{c|}{Chamfer Dist \( \downarrow \)} & \multicolumn{2}{c|}{Volume IoU \( \uparrow \)} \\
        \cline{2-3} \cline{4-5}
        & \cite{Nicolet2021Large} & \textbf{Ours} & \cite{Nicolet2021Large} & \textbf{Ours} \\
        \hline
        Kitten &0.0031  &\textbf{0.0020}  &0.2673  &\textbf{0.9192}  \\
        Amphora &0.0032  &\textbf{0.0021}  &0.8171  &\textbf{0.8951} \\
        Pretzel &0.0049  &\textbf{0.0043}  &0.7041  &\textbf{0.7743}  \\
        Birthday &0.0021  &\textbf{0.0008}  &0.5083  &\textbf{0.7706}  \\
        Elephant &0.0035  &\textbf{0.0020}  &0.5297  &\textbf{0.6288}  \\
        Botijo &0.0053  &\textbf{0.0030}  &0.4639  &\textbf{0.8159}  \\
        Sorter &0.0173  &\textbf{0.0032}  &0.6142  &\textbf{0.7464}  \\
        Heptoroid &0.0074  &\textbf{0.0062}  &0.5642  &\textbf{0.7272}  \\
        \hline
    \end{tabular}
    \label{tab:comparison}
\end{table}

\section{Conclusion}
In this paper, we proposed a collaborative inverse rendering strategy that leverages persistent homology priors jointly with the uniform-sphere assumption to provide a principled basis for camera placement. Compared to the state-of-the-art neural network-free mesh-based inverse rendering method of Nicolet et al.~\cite{Nicolet2021Large}, our approach achieves superior reconstruction quality on challenging high-genus models. By guiding camera placement through topological invariants such as tunnels, persistent homology priors enhance robustness against gradient vanishing, occlusion, and topological failure. These results demonstrate that incorporating topological priors into mesh-based inverse rendering offers a promising direction for accurate and robust high-genus surface reconstruction. Looking forward, we aim to extend this strategy to non-mesh-based, deep learning-based image-to-3D tasks to better address challenging high-genus structures.

\vfill\pagebreak

\clearpage

\bibliographystyle{IEEEbib}
\bibliography{refs}

@String(TOG= {ACM Trans. Graph.})

@String(ICLR = {Int. Conf. Learn. Represent.})

@String(TOG   = {ACM TOG})

@String(ICLR  = {ICLR})

@article{Nicolet2021Large,
  author = {Baptiste Nicolet and Alec Jacobson and Wenzel Jakob},
  title = {Large Steps in Inverse Rendering of Geometry},
  journal = {ACM Transactions on Graphics (TOG)},
  volume = {40},
  number = {6},
  article = {248},
  year = {2021},
  month = {December},
  pages = {13},
  doi = {10.1145/3478513.3480501}
}

@InProceedings{KingBa15,
  author    = {Kingma, Diederik and Ba, Jimmy},
  booktitle = {International Conference on Learning Representations (ICLR)},
  title     = {Adam: A Method for Stochastic Optimization},
  year      = {2015},
  address   = {San Diega, CA, USA},
  optmonth  = {12},
}

@misc{liu2023zero1to3,
      title={Zero-1-to-3: Zero-shot One Image to 3D Object}, 
      author={Ruoshi Liu and Rundi Wu and Basile Van Hoorick and Pavel Tokmakov and Sergey Zakharov and Carl Vondrick},
      year={2023},
      eprint={2303.11328},
      archivePrefix={arXiv},
      primaryClass={cs.CV}
}

@article{liu2023syncdreamer,
  title={SyncDreamer: Generating Multiview-consistent Images from a Single-view Image},
  author={Liu, Yuan and Lin, Cheng and Zeng, Zijiao and Long, Xiaoxiao and Liu, Lingjie and Komura, Taku and Wang, Wenping},
  journal={arXiv preprint arXiv:2309.03453},
  year={2023}
}

@article{guo2024tetsphere,
  title={TetSphere Splatting: Representing High-Quality Geometry with Lagrangian Volumetric Meshes},
  author={Guo, Minghao and Wang, Bohan and He, Kaiming and Matusik, Wojciech},
  journal={arXiv preprint arXiv:2405.20283},
  year={2024}
}

@article{long2023wonder3d,
  title={Wonder3D: Single Image to 3D using Cross-Domain Diffusion},
  author={Long, Xiaoxiao and Guo, Yuan-Chen and Lin, Cheng and Liu, Yuan and Dou, Zhiyang and Liu, Lingjie and Ma, Yuexin and Zhang, Song-Hai and Habermann, Marc and Theobalt, Christian and others},
  journal={arXiv preprint arXiv:2310.15008},
  year={2023}
}

@article{Botsch2004a,
  author = {Mario Botsch and Leif Kobbelt},
  title = {An intuitive framework for real-time freeform modeling},
  journal = {ACM Transactions on Graphics},
  volume = {23},
  number = {3},
  year = {2004},
  pages = {630--634},
  doi = {10.1145/1015706.1015772}
}

@article{Jacobson2010,
  author = {Alec Jacobson and Elif Tosun and Olga Sorkine and Denis Zorin},
  title = {Mixed Finite Elements for Variational Surface Modeling},
  journal = {Computer Graphics Forum},
  volume = {29},
  number = {5},
  pages = {1565--1574},
  year = {2010},
  doi = {10.1111/j.1467-8659.2010.01765.x}
}

@article{downs2022google,
  title={Google Scanned Objects: A High-Quality Dataset of 3D Scanned Household Items},
  author={Downs, Laura and Francis, Anthony and Koenig, Nate and Kinman, Brandon and Hickman, Ryan and Reymann, Krista and McHugh, Thomas B and Vanhoucke, Vincent},
  journal={arXiv preprint arXiv:2204.11918},
  year={2022}
}

@inproceedings{Jung2023,
  author = {Yucheol Jung and Hyomin Kim and Gyeongha Hwang and Seung-Hwan Baek and Seungyong Lee},
  title = {Mesh Density Adaptation for Template-based Shape Reconstruction},
  year = {2023},
  doi = {10.1145/3588432.3591498},
  booktitle = {ACM SIGGRAPH 2023 Conference Proceedings},
  publisher = {ACM},
  url = {https://doi.org/10.1145/3588432.3591498},
  location = {Los Angeles, CA, USA}
}

@software{Mitsuba3,
    title = {Mitsuba 3 renderer},
    author = {Wenzel Jakob and Sébastien Speierer and Nicolas Roussel and Merlin Nimier-David and Delio Vicini and Tizian Zeltner and Baptiste Nicolet and Miguel Crespo and Vincent Leroy and Ziyi Zhang},
    note = {https://mitsuba-renderer.org},
    version = {3.1.1},
    year = 2022
}

@article{Edelsbrunner2002,
  author    = {Herbert Edelsbrunner and David Letscher and Afra Zomorodian},
  title     = {Topological persistence and simplification},
  journal   = {Discrete \& Computational Geometry},
  year      = {2002},
}

@book{EdelsbrunnerHarer2022,
  author    = {Herbert Edelsbrunner and John L. Harer},
  title     = {Computational Topology: An Introduction},
  publisher = {American Mathematical Society},
  year      = {2022},
}

@article{ghrist2008barcodes,
  title={Barcodes: the persistent topology of data},
  author={Ghrist, Robert},
  journal={Bulletin of the American Mathematical Society},
  volume={45},
  number={1},
  pages={61--75},
  year={2008}
}

@article{PH,
author = {Edelsbrunner, Herbert and Harer, John},
year = {2008},
month = {01},
pages = {},
title = {Persistent homology—a survey},
volume = {453},
isbn = {9780821842393},
journal = {Discrete \& Computational Geometry - DCG},
doi = {10.1090/conm/453/08802}
}

@article{Thingi10K,
  title={Thingi10K: A Dataset of 10,000 3D-Printing Models},
  author={Zhou, Qingnan and Jacobson, Alec},
  journal={arXiv preprint arXiv:1605.04797},
  year={2016}
}

@misc{gao2025inverserenderinghighgenussurface,
      title={Inverse Rendering for High-Genus Surface Meshes from Multi-View Images}, 
      author={Xiang Gao and Xinmu Wang and Xiaolong Wu and Jiazhi Li and Jingyu Shi and Yu Guo and Yuanpeng Liu and Xiyun Song and Heather Yu and Zongfang Lin and Xianfeng David Gu},
      year={2025},
      eprint={2511.18680},
      archivePrefix={arXiv},
      primaryClass={cs.GR},
      url={https://arxiv.org/abs/2511.18680}, 
}

@misc{gao2025neural,
      title={Neural Geometry Image-Based Representations with Optimal Transport (OT)}, 
      author={Xiang Gao and Yuanpeng Liu and Xinmu Wang and Jiazhi Li and Minghao Guo and Yu Guo and Xiyun Song and Heather Yu and Zhiqiang Lao and Xianfeng David Gu},
      year={2025},
      eprint={2511.18679},
      archivePrefix={arXiv},
      primaryClass={cs.CV},
      url={https://arxiv.org/abs/2511.18679}, 
}

@inproceedings{otTalk,
author = {Wang, Xinmu and Gao, Xiang and Song, Xiyun and Yu, Heather and Lin, Zongfang and Peng, Liang and Gu, Xianfeng},
title = {OT-Talk: Animating 3D Talking Head with Optimal Transportation},
year = {2025},
isbn = {9798400718779},
publisher = {Association for Computing Machinery},
address = {New York, NY, USA},
url = {https://doi.org/10.1145/3731715.3733411},
doi = {10.1145/3731715.3733411},
booktitle = {Proceedings of the 2025 International Conference on Multimedia Retrieval},
pages = {1340–1349},
numpages = {10},
location = {Chicago, IL, USA},
series = {ICMR '25}
}

\end{document}